\documentclass[sn-mathphys-num]{sn-jnl}

\usepackage{graphicx}%
\usepackage{multirow}%
\usepackage{amsmath,amssymb,amsfonts}%
\usepackage{amsthm}%
\usepackage{mathrsfs}%
\usepackage[title]{appendix}%
\usepackage{xcolor}%
\usepackage{textcomp}%
\usepackage{manyfoot}%
\usepackage{booktabs}%
\usepackage{algorithm}%
\usepackage{algorithmicx}%
\usepackage{algpseudocode}%
\usepackage{listings}%
\usepackage{todonotes}
\usepackage{longtable}
\usepackage{ltablex}     
\keepXColumns            

\usepackage{enumitem}  
\usepackage{amssymb}   

\begin{document}

\title[Automated Scientific Discovery]{Automated Scientific Discovery: From Equation Discovery to Autonomous Discovery Systems }


\author*[1]{\fnm{Stefan} \sur{Kramer}}\email{kramer@informatik.uni-mainz.de}

\author[1]{\fnm{Mattia} \sur{Cerrato}}\email{mcerrato@uni-mainz.de}

\author[2]{\fnm{Jannis} \sur{Brugger}}\email{jannis.brugger@tu-darmstadt.de}

\author[3]{\fnm{Sašo} \sur{Džeroski}}\email{Saso.Dzeroski@ijs.si}

\author[4,5]{\fnm{Ross D.} \sur{King}}\email{rk663@cam.ac.uk}

\affil*[1]{\orgdiv{Computer Science Department}, \orgname{Johannes Gutenberg University Mainz}, \orgaddress{\street{Saarstrasse 21}, \city{Mainz}, \postcode{55116}, \country{Germany}}}

\affil[2]{\orgdiv{hessian.AI}, \orgname{TU Darmstadt}, \orgaddress{\street{Karolinenpl. 5}, \city{Darmstadt}, \postcode{64289}, \country{Germany}}}

\affil[3]{\orgdiv{Dept. of Knowledge Technologies}, \orgname{Jozef Stefan Institute}, \orgaddress{\street{Jamova cesta 39}, \city{Ljubljana}, \postcode{1000}, \country{Slovenia}}}

\affil[4]{\orgdiv{Data Science and AI}, \orgname{Chalmers University of Technology}, \orgaddress{\street{Chalmersgatan 4}, \city{Göteborg}, \postcode{41296}, \country{Sweden}}}

\affil[5]{\orgdiv{Department of Chemical Engineering and Biotechnology}, \orgname{University of Cambridge}, \orgaddress{\street{Philippa Fawcett Drive}, \city{Cambridge West}, \postcode{CB3 0AS
}, \country{United Kingdom}}}


\abstract{The paper surveys automated scientific discovery, from equation discovery and symbolic regression to autonomous discovery systems and agents. It discusses the individual approaches from a "big picture" perspective and in context, but also discusses open issues and recent topics like the various roles of deep neural networks in this area, aiding in the discovery of human-interpretable knowledge. Further, we will present closed-loop scientific discovery systems, starting with the pioneering work on the Adam system up to current efforts in fields from material science to astronomy. Finally, we will elaborate on autonomy from a machine learning perspective, but also in analogy to the autonomy levels in autonomous driving. The maximal level, level five, is defined to require no human intervention at all in the production of scientific knowledge. Achieving this is one step towards solving the Nobel Turing Grand Challenge to develop AI Scientists: AI systems capable of making Nobel-quality scientific discoveries highly autonomously at a level comparable, and possibly superior, to the best human scientists by 2050.}

\maketitle

\section{Introduction}

The automated discovery of scientific knowledge has always been on the agenda of artificial
intelligence research, and prominently so since the end of the 1970s \citep{Langley1977, Langley1987}.
Scientific knowledge takes many forms: In many cases, the scientific process begins with collecting
and classifying objects, and creating taxonomies of classes of objects. The more a scientific
discipline advances, the more it tends to strive to describe the phenomena quantitatively, for
better explanation and prediction. By far the most commonly used representation for describing
systems of interest is in the form of mathematical equations, in particular differential equations. 
Thus, the automated discovery of equations from data has been established as a family of methods
within and partly outside artificial intelligence: it runs under the heading of equation discovery
\citep{Langley1977,Dzeroski1993} as well as symbolic regression \citep{Koza1994}.

The goal in many application domains of equation discovery and symbolic regression is to learn
a human-understandable model of the system dynamics in the form of (mostly ordinary) differential
equations.\footnote{The underlying data are most frequently temporal.} One important aspect of
scientific discovery is that the resulting models need to be in principle interpretable.\footnote{If a
model cannot be communicated to a community of researchers, it hardly qualifies as scientific,
as communication is an indispensable part of the scientific endeavor.} Thus, the goal is not
optimization (e.g., of properties in material science or drug development), but to develop understanding.

An important part of the literature on automated scientific discovery \citep{Langley1987,Li2021}
discusses the topic from a cognitive science point of view (what are or could be the reasoning
processes leading to certain discoveries?) and thus also a historical reconstruction of the processes.
This is relevant, because today's AIs for scientific discovery also have to start from the same
principles to enable discoveries in completely new application domains. While this can be viewed
on the symbolic level only, many of today's approaches also consider the subsymbolic level to aid
the process: neural networks of various sorts can play a vital role in guiding the search, providing
valuable information to the discovery agent, or turning low-level sensory information into high-level
information that can be used for symbolic reasoning.

Finally, the question of autonomy of the discovery agents arises. While early systems assumed
a table of input data is given by a human user, approaches with more autonomy on the side of
the discovery agent are becoming more common. The approach became prominent with the development
of the first robot scientist world-wide, Adam \citep{King2009}, that automated cycles of hypothesis
generation and testing in the field of functional genomics. Meanwhile, the third generation of
robot scientists is being developed. The degrees of autonomy of a discovery agent may range from
completely passive, i.e., supervised learning, via active learning \citep{Cohn1996} to reinforcement
learning \citep{Sutton2018}.

\begin{figure}[t!]
    \centering
    \includegraphics[width=1.0\linewidth]{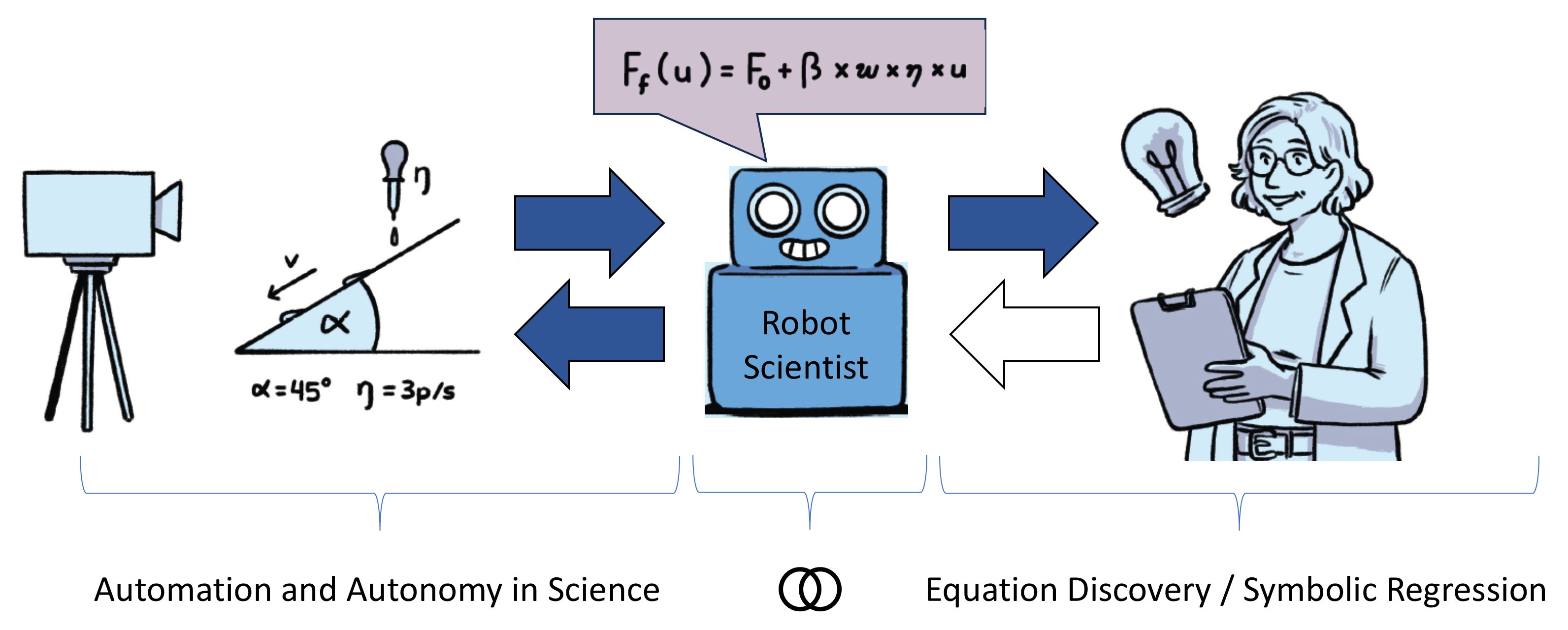}
    \caption{Overview of the two realms of automated scientific discovery: (i) the discovery and communication of human-interpretable knowledge in a representation used by scientists in the field, e.g., equations (right-hand side) and (ii) autonomy and automation in science (left-hand slide). Approaches integrating both are currently rare.}
    \label{fig:scientific-discovery}
\end{figure}

Considering the above, this paper aims to give an overview of automated scientific discovery from a conceptual
point of view, spanning the whole field from the generation of scientific knowledge, mainly in the form of
equations, to automation and autonony in robot scientists or self-driving labs. It does not just enumerate 
approaches, but discusses central conceptual aspects and open issues that need to addressed in 
future systems. Particular attention is paid to the role of neural networks in the process: either
for representation learning, for search in neural-guided equation discovery, or in neural operators,
which abandon interpretability altogether. Discussing two main aspects of automated scientific discovery
side by side in one paper, (i) the discovery of interpretable scientific knowledge in the form of 
equation on the one hand and (ii) automation/autonomy on the other (see Figure \ref{fig:scientific-discovery}),
we identify a major research gap: systems that run autonomously, but are able to communicate
results in formalisms used by scientists, so that interventions are possible, such as hints for search,
the provision of goals and values, and the embedding of findings in bigger theories. Very few
systems exist in this space, however, we would like mention the pioneering work of Jan Zytkow,
who coupled real electrochemistry experiments with the FAHRENHEIT system for equation
discovery \cite{ZytkowAAAI1990}, and later proposed a robotic system for the rediscovery
of Galileo's equation of objects rolling down an inclined plane, again with the help of FAHRENHEIT,
but already taking into account empirical error \cite{ZytkowISMIS1997}.

The paper is structured as follows: In Section~2, we will review equation discovery and
symbolic regression from the beginnings to the current state of the art, with a list of open problems.
In Section~3, we discuss the representations used in current scientific discovery and, in particular,
how neural networks can be employed to learn representations for the discovery process and how
dynamics can be learned directly by neural operators. The topic of Section~4 is closed-loop
scientific discovery, with recent progress in the field. Section~5 discusses different levels of autonomy.
An overview of benchmarks and testbeds is given in Section~6, before we conclude in Section~7.

The survey paper is different from existing papers in many ways: Makke and Chawla \cite{ChawlaAIR2024}
presented a thorough survey of symbolic regression (SR) and equation discovery (ED). Our survey covers
{\em both} SR/ED and automation/autonomy, so it is broader in scope. Also, it appears more conceptual and
with a stronger focus on interesting open issues. Further, in the present paper the discussion of the various uses
of neural networks appears both more extensive and deeper.  In a recent study, Musslick {\em et al.}
\cite{MusslickPNAS2025} discuss primarily the limitations of autmated scientific discovery, with a focus
on societal and ethical implications (e.g., the value alignment of robot scientists with human scientists).
It discusses what should not be done, but also what potentially cannot be done. The latter is, of course,
harder to argue, as it involves a forecast of the further progress of the field of artificial intelligence
in general. Arguments likes the paradox of automation hold, others concerning the computational complexity
of scientific discovery require more investigation. Another recent survey by Gao {\em et al.}  \cite{ZitnikCell2024}
focuses on life sciences exclusively and discusses everything in terms of agentic AI, which is both not
our emphasis here.  Two recent papers by Pat Langley \citep{LangleyAIMag2022, LangleyAAAI2024} are both
related, but at the same time different.  The first of them \cite{LangleyAIMag2022} discusses the
so far distinct notions of ``agents of exploration'' and ``agents of discocery''.  Langley argues for a 
synthesis of the two, such that agents can both explore and discover in remote areas, like in
space or in the deep sea. Although conceptually relevant (imagine a versatile scientific agent that
explores a lab environment and discovers new concepts and laws along the way), the main thrust
of the paper is clearly different. In the more recent paper \cite{LangleyAAAI2024}, Langley describes
an integration effort different from the one shown above: 
In the paper, he envisages a tight integration and coupling of the various phases of scientific
discovery, from the discovery of taxonomic knowledge via qualitative models to quantitative and
causal models.  It is argued that this integration is important, but has not been achieved before.
We believe that, while interresting, this is of a different nature than the integration between the
discovery of scientific, human-interpretable knowledge, and automation and autonomy in robot
scientists or self-driving labs (see Figure \ref{fig:scientific-discovery}).


\section{From BACON to Modern Equation Discovery and Symbolic Regression}
\subsection{History and Current Approaches}

\begin{figure}[t!]
    \centering
    \includegraphics[width=0.9\linewidth]{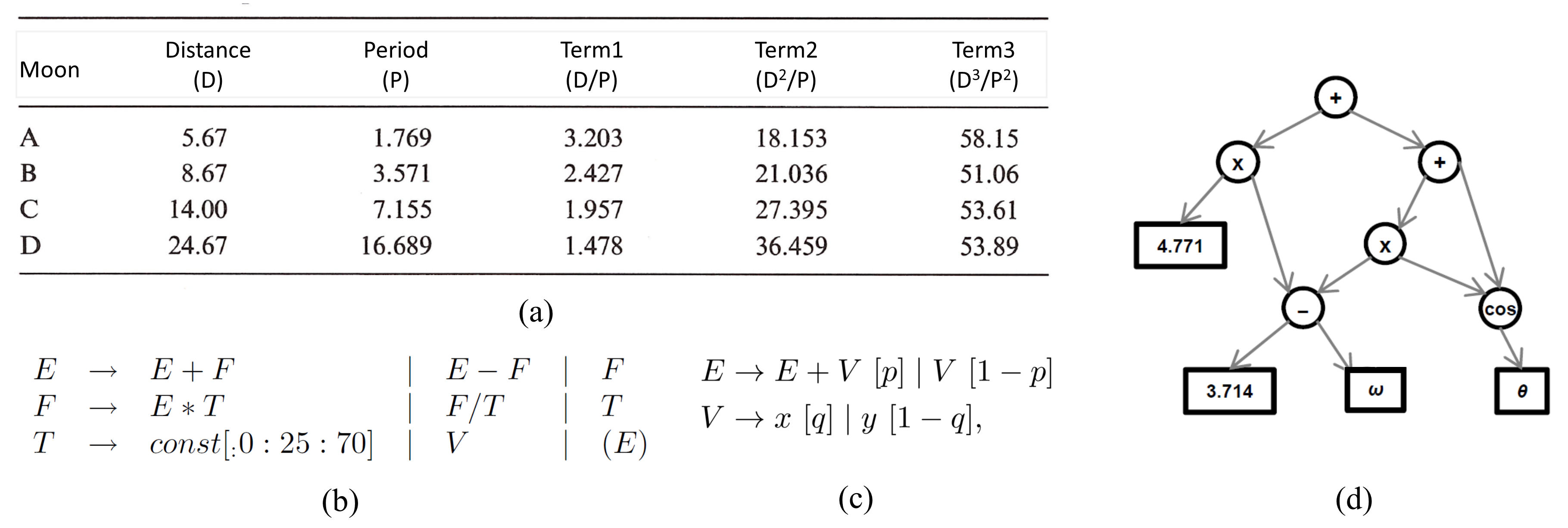}
    \caption{(a) BACON \cite{Langley1977,Langley1987} (b) Example of context-free grammar guiding the search for equations in the Lagramge system \cite{Todorovski1997} (c) A probabilistic context-free grammar as used in ProGED \cite{Brence2021} (d) Symbolic regression \cite{Schmidt2009}}
    \label{fig:eq-discovery}
\end{figure}

The first system for the discovery of equations based on data was BACON by Pat Langley \citep{Langley1977}, represented in Figure~\ref{fig:eq-discovery}. The first version of BACON was developed into a series of following systems, BACON.2 to BACON.5, with increasingly complex functionality \citep{Langley1987}. The basic philosophy behind the book by Langley et al. was that scientific discovery, even in its most intricate ways, is essentially problem solving. This even applies to the search for new problems, new representations, and new measurement devices. In the case of the BACON systems, the idea was applied to the discovery of equations.

BACON.1 to BACON.5 were implemented on the basis of PRISM, a system for the representation and inference of production rules. The BACON systems relied on the observation of the correlation of pairs of variables, when everything else is being held constant (ceteris paribus). This is a strong assumption, as it will in many cases not be possible to control all other variables in an experiment. Also, interestingly, BACON has a flavor of active learning, since users are requested to record data, if they are not available yet. One interesting feature of BACON is the construction of new terms, e.g., ratios or products of existing terms, by production rules. In this way, it takes advantage of the structure of the search space, which is rarely ever attempted in current systems. Noise handling is achieved by some tolerance parameter, which establishes that a value of a variable (constructed or initially given) is constant, even though it varies within a certain range. BACON.2 to BACON.5 included advanced features for symmetries, common divisors, and conservation laws, amongst others. Fig.~1(a) shows the derivation of Kepler's third law $D^3/P^2 = k$ by a sequence of newly constructed terms, until a --- more or less --- constant value is found.

The next generation of equation discovery systems was not restricted to keeping all but a pair of variables fixed, but was able to handle observational data. In addition, it was able to learn models of dynamical systems in the form of ordinary differential equations (ODEs). Lagrange \citep{Dzeroski1993} computes all derivatives up to a pre-defined order, then generates products of up to a maximum of variables, before it calculates a simple linear regression to generate a candidate equation. More recently, this approach has been taken up in the SINDY system \citep{Brunton2016}, which applies sparse (instead of simple) linear regression. 
In the meantime, the method has been extended to capture nonlinear dynamics by shallow recurrent decoder networks (SINDy-SHRED) \cite{gao2025sparseidentificationnonlineardynamics}.
The successor of Lagrange, named Lagramge \citep{Todorovski1997}, was a milestone in equation discovery, as it introduced the use of domain knowledge in addition to data: It was the first system to use a context-free grammar (CFG) to guide the search for systems of equations. Grammars are a way for domain experts to use prior knowledge and let that knowledge guide the search for equations. In this way, Lagramge was able to solve problems that the predecessor Lagrange was not able to solve, for instance, the problem of two poles on a cart. An example CFG for Lagramge is shown in Figure~1(b). Lagramge GSAT \citep{Ganzert2010} improves Lagramge by a bundle of modifications: first, it uses a search procedure similar to GSAT (random restart hillclimbing) to randomize search; further, it employs a one-step look-ahead and a momentum to make the search less erratic. Washio \& Motoda \citep{Washio1997} further improved the methods by also taking into account units and scale types as constraints. Dimensional units are also considered for use in ProGED \citep{Brence2021,Brence2023}, which is based on the idea of using probabilistic CFGs to represent the search space and sample from it. An example is given in Fig.~1(c), where both the rules and the probabilities associated with the rules (p and q) are shown. These probabilities can be fixed, but can also be learned from corpora of equations \citep{Chaushevska2022}. Sampling candidate equations from probabilistic CFGs enables easy parallelization: batches of sampled equations can be distributed to nodes and evaluated in an embarrassingly parallel way.

Symbolic regression, a development parallel to the development of equation discovery, was originally based on genetic programming (GP): the term was introduced by Koza \citep{Koza1994}. Typical systems of symbolic regression work on an operator tree or DAG representation of equations (see Fig.~1(d)). These trees are modified by a set of possible operations, such as crossover between subtrees of two trees (equations), mutations, substitutions of variables by constants, or, vice versa, substitutions of constants by variables. Schmidt \& Lipson \citep{Schmidt2009} used symbolic regression to discover natural laws from measured data. Symbolic regression approaches were used early on to discover ODEs \citep{Dzeroski1994} and used ideas from grammar-based genetic programming to consider domain-specific knowledge, paving the way for systems that use both data and domain knowledge in equation discovery, such as Lagramge, Lagramge2.0 \citep{Todorovski2006}, IPM \citep{Bridewell2008} and Prob-MoT \citep{Cherepnalkoski2012}. The last three use process-based formalism to represent models and domain knowledge.

The Bayesian machine scientist \citep{Guimera2020} establishes the plausibility of models using explicit approximations to the exact marginal posterior over models and establishes its prior expectations about models by learning from a large empirical set of mathematical expressions. The space of equations is explored via Markov Chain Monte Carlo (MCMC), with specific moves for mathematical expression sampling.

PySR  \cite{CranmerPySR} is a fast, effective and popular approach to symbolic regression. It is based on genetic programming and outputs one solution per complexity class (from simple to complex equations). PySR is frequently found to be well-performing in practice. It has a Python front-end and delegates numerical computations to Julia. Using Julia ``under the hood'' and heuristics to avoid redundancy, it is able to explore a large number of candidates in a relatively short period of time, giving it a competitive advantage in many situations. In the meantime, version 1.4 of PySR is available with template expressions and version 1.5 with mini-batching, which further improves practical applicability.

Recent work by Boris Kr\"amer and collaborators \cite{KraemerSIAM2024} has advanced the use of quadratic models for data-driven discovery of dynamical systems governed by partial differential equations (PDEs). In particular, they explore transformations of nonlinear PDEs into quadratic form, which enables the application of structure-preserving reduced-order modeling and symbolic regression techniques. The approach facilitates the use of quadratic latent variable models that retain interpretability and allow for efficient training on noisy and sparse data. The usefulness of the approach has been demonstrated in areas such as fluid dynamics and plasma modeling.

Symbolic regression and equation discovery are currently limited to systems with only few variables. Xue et al. \cite{XueECML2023} address this problem by identifying control variables, which can be varied to discover the dynamics of a system in ``controlled experiments'' step by step. The approach is still based on genetic programming. A precondition of its use is evidently the existence of such variables, which is not always the case. In practical applications and real systems, the set of control variables is not equal to the set of variables that should appear in an equation. Thus, that mapping has to be learned first. Nevertheless, the idea of actively using control variables to reduce complexity is valuable and could be a key to making ED/SR practically applicable to large and complex sytems.    


In recent years, a new field of research has emerged that focuses on how neural networks can be used in equation discovery. To provide an overview, we divide the works into three categories. The categories are:
(i) NN as a supporting module in the equation discovery system (EDS), (ii) NN as the main component of the EDS, and finally, (iii) foundation models as EDS. We discuss the three categories in consecutive order.

AI Feynman 2.0  \citep{Udrescu2020} is a recent symbolic regression approach that aims to improve its predecessor (a) by structuring the search space by building the equation in meaningful increments and (b) making it more noise-tolerant. The first goal is achieved by graph modularity, i.e., constructing the equations by so-called graph modules. It should be noted that, in doing so, it is one of the few approaches that takes advantage of the structure of the search space (instead of just brute-force search, sampling or ``blind'' randomized traversal). The second goal is achieved by employing an MDL-inspired evaluation function instead of the RMSE. This function is called MEDL in Feynman 2.0. Using MEDL, effective pruning can be developed, because the complexity of the equation can be balanced against its error.
Lusch et al. \cite{lusch2018deep} apply an auto-encoder structure to find a coordination transformation for a differential equation that maps the nonlinear original problem to linear embedding.
Following the idea of an autencoder, Mežnar et al. \cite{mevznar2023efficient}  embed equations in a low-dimensional latent space and use this smooth latent space to suggest new equations based on genetic programming.
Mundhenk et al. \cite{mundhenk2021symbolic} use a Recurrent Neural Network (RNN) to seed a genetic programming algorithm, and the genetic algorithm results are used to train the RNN. While the previous works use a subsymbolic component to simplify the original problems, the following articles use neural networks as main component.

Deep Symbolic Regression (DSR) \citep{Petersen2021} addresses the problem of GP approaches with finding solutions for larger problems. It employs a recurrent neural network to build an equation tree step by step. As the objective function (of fitting a low-error equation) is not differentiable, a reinforcement learning approach is proposed. More specifically, DSR employs a risk-seeking policy gradient, which maximizes the best-case performance, not the average-case performance. \emph{NeSymReS} \cite{biggio_neural_2021}, \emph{SymbolicGPT}  \cite{valipour_symbolicgpt_2021}, and  \emph{E2E} \cite{kamienny_end--end_2022}, use a transformer-based architecture to predict the equation on a token level. The main difference is the embedding architecture of the data set. \emph{MGMT} \cite{brugger2025neuralguidedequationdiscovery} compares different embedding methods and shows their influence on the prior of the guiding network. Additionally, the work shows that supervised learning is beneficial compared to reinforcement learning for the architectures considered. \emph{TPSR} \cite{shojaee_transformer-based_2023} and \emph{DGSR-MCTS} \cite{kamienny_deep_2023}, combine a transformer-based architecture with a Monte Carlo Tree Search (MCTS). In the second paper, the network suggests how to mutate the current equation.
Another approach is to train a specialized end-to-end differentiable network and parser it after the training with gradient descent to an equation. \emph{EQL}$^{\div}$ \cite{sahoo_learning_2018} or Kolmogorov Arnold networks (KAN) \cite{liu2024kan} are examples for this approach. 

Large language models (LLMs) have also impacted the field of equation discovery. Foundation models such as GPT-4 have the advantage that after the initial learning, they only need to be adapted to the equation discovery domain through fine-tuning or prompt design. In addition, they have been shown to retain background knowledge from the initial training, and the user can add domain knowledge through prompts.
In-Context Symbolic Regression \emph{ICSR} \cite{merler-context_2024}, and Sharlin et al. \cite{sharlin_context_2024} employ a foundation model to produce initial equations. These equations are then tested on the data set. The fitness score and other measures, such as complexity, are calculated externally and then fed back to the model with the task of refining the solutions.
\emph{LLM-SR} \cite{shojaee_llm-sr_2024} follows the same idea but represents equations as programs and uses comments in the program to make the discovered equation better understandable.
Meyerson et al. \cite{meyerson_language_2023} use a foundation model to perform genetic programming (mutation, crossover, etc.) through prompts.
The foundation model-based equation discovery systems show promising results, but the extent to which the initial training influences the test results has not yet been sufficiently investigated, as the standard benchmarks (see below) are included in the initial training.

\subsection{Open Problems}
In equation discovery and symbolic regression, a few open problems can be identified:
\begin{itemize}
    \item It remains hard to exploit structure in the space of equations to guide the search to promising parts of the search space. Opportunities for pruning would also be helpful.
    \item At least in the case of differential equations, fitting the model is the most expensive part. Ways of stopping the fitting process if it turns out to be unpromising would save a lot of computation time.
    \item Equations are ``brittle'': properties of differential equations can change dramatically with only little syntactic modifications. Minor changes can lead to no solutions, one solution, or many solutions.
    \item Most approaches struggle with a dimensionality of the problem higher than a very small number of variables. 
    \item Overfitting avoidance and regularization: The syntactic complexity of an equation does not necessarily correspond to the complexity of the function in the feature space. Meaningful ways to approximate or bound complexity would be helpful.
    \item For the approaches based on foundation models, it is unclear how the results can generalize to new, previously unseen problems. Data provenance is an issue: It is unclear whether the models have seen some of the equations before in training. Many of the approaches are based on embeddings of datasets. It is, at this point, not clear, what the best way is to embed a dataset for a foundation model for symbolic regression.
    \item Relating discovered equations to existing theory or making the equations consistent with it remains a big challenge. Quite related, it is not clear whether or how ``understanding of the physical meaning'' of variables can be achieved.
\end{itemize}

\section{Representation Learning in Scientific Discovery}
\label{sec:representationlearning}

\subsection{Representation Learning of the Input}

The standard representation of data for scientific discovery is tabular data (see, e.g., also the tables in the book by Langley et al. \citep{Langley1987} and Figure~1(a)). However, recent years have seen a surge of papers that use neural networks as an intermediate representation to aid in the discovery of models.

One notable example is the work of Miles Cranmer and Shirley Ho \citep{Cranmer2020}, who proposed Graph Neural Networks (GNNs) as an intermediate representation. GNNs were used to learn about the interaction of objects, in terms of, for example, forces that act upon each other. Classical examples include n-body problems or, more specifically, orbital mechanics---the motion of planets and other larger objects in our solar system. The nodes in the graph represent the objects, which are annotated by feature vectors representing the properties of the objects. The edges in the graph represent the interactions between the objects and are annotated by properties that partially depend on those of the objects. For example, one may consider the masses of planets as properties of the nodes, and the distance and gravitational force between the objects as properties of the edges. When learning GNNs, typically, so-called node models $\phi_v$ are updated depending on the edge models $\phi_e$ of neighboring edges and, alternatingly, the edge models $\phi_e$ are updated based on the node models $\phi_n$ of the nodes that the edges connect. Update steps are frequently framed as message passing, and pooling functions aggregate input from multiple edges connected to one node. GNNs usually can be trained end-to-end, but are not guaranteed to converge.

\begin{figure}[t!]
    \centering
    \includegraphics[width=0.7\linewidth]{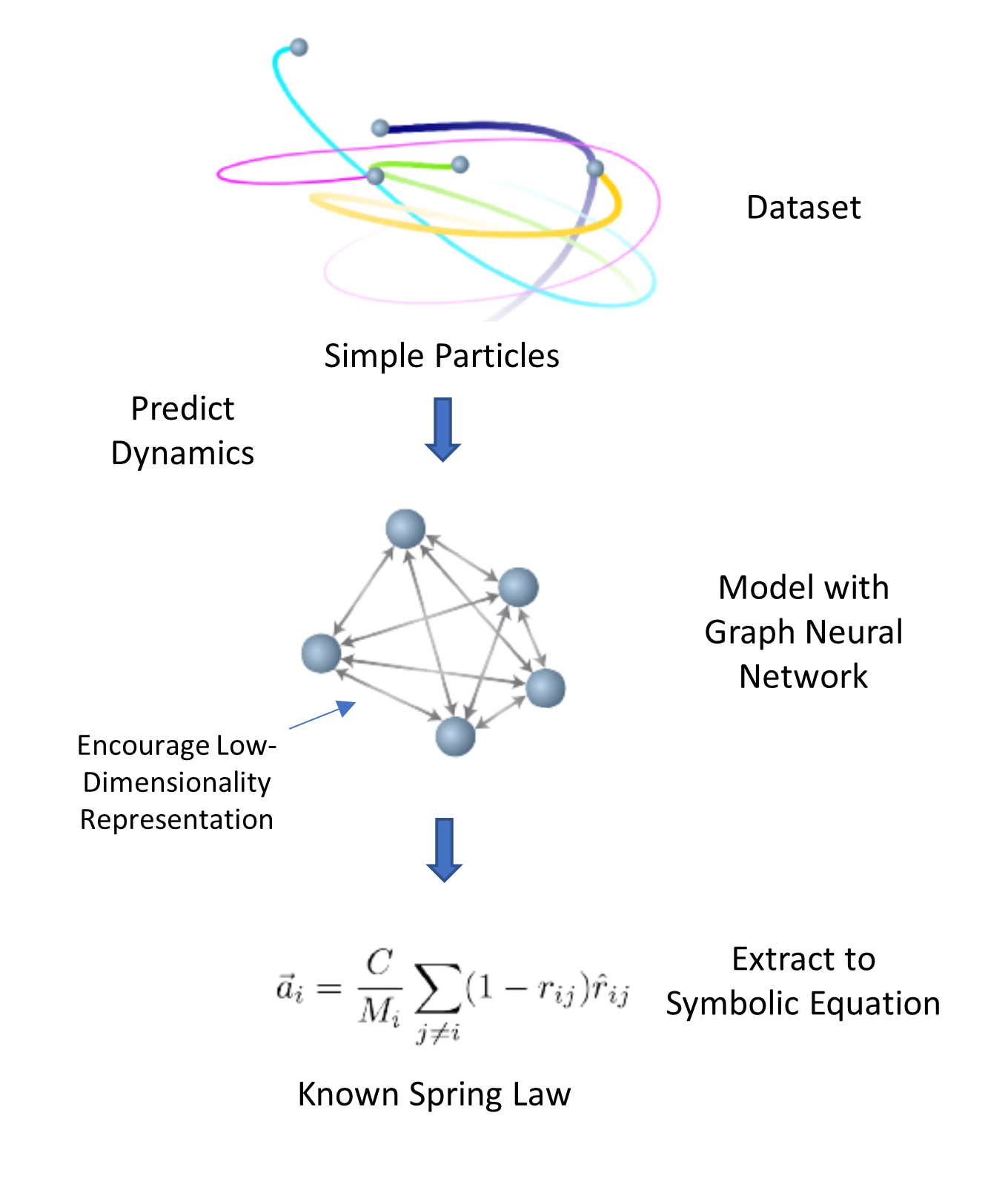}
    \caption{Workflow of Cranmer et al. \cite{Cranmer2020}: GNNs as an intermediate representation to support or enable the learning process}
    \label{fig:cranmer}
\end{figure}

In the application domain that was given as an example, orbital mechanics, the input to the system are $(x,y,z)$ coordinates of the Sun, all planets, and all moons with a mass above $10^{18}$~kg. Data from 1980 to 2013 were used with time intervals of 30 minutes each, with the first 30 years for training and the last 3 years for validation.

\begin{figure}[h]
    \centering
    \includegraphics[width=0.6\linewidth]{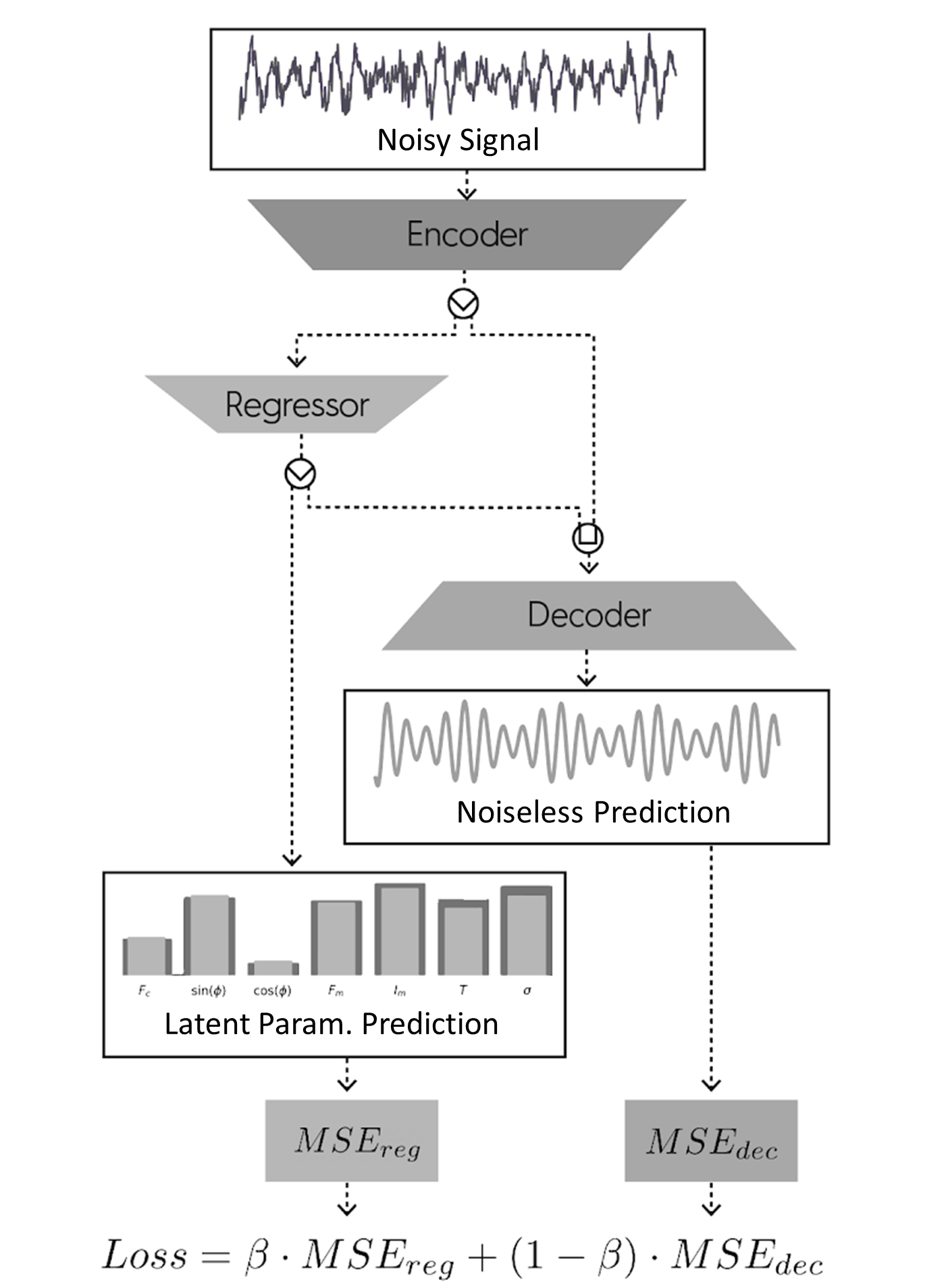}
    \caption{Neural network architecture of model that extracts known and unknown physical parameters from oscillating time series \cite{Garcon2022}.}
    \label{fig:garcon}
\end{figure}

Garcon et al. \citep{Garcon2022} proposed a method to predict known physical parameters and discover new ones from oscillating time series (Figure~\ref{fig:garcon}). The method is trained on a large set of synthetic time series. The latent parameters used to generate the monochromatic sine waves are the carrier frequency, $F_c$, and phase $\phi$ (which is mapped for technical reasons to two separate parameters, $\sin(\phi)$ and $\cos(\phi)$), in addition to the coherence time $\tau$. The AM and FM sine waves are generated by adding a modulation function to the carrier. The modulation function's latent parameters are the modulation frequency $F_m$ and amplitude $I_m$. Noise is linearly added to the pure signals by sampling the Gaussian distribution. AM/FM signals with minimum $I_m$ reduce to decaying monochromatic sine waves and reach 100\% modulation with maximum $I_m$. These latent parameter ranges are wide enough such that they would encompass many foreseeable real-world signals. Figure~3 shows the neural network architecture used to predict the latent parameters, with an autoencoder-type subnetwork to support the prediction. The method can be used to discover new parameters (not just predict known ones) and reconstruct equations producing input time series.

The situation is clearly more complex when the observations are given as videos instead of tabular data. Chen et al. \citep{Chen2022} presented a solution based on what they call neural state variables. Neural state variables are essentially latent variables. The current state-of-the-approach to computing latent variables is to define an autoencoder with a bottleneck layer of the right dimension. The dimension should be large enough to allow faithful reconstruction by the decoder, but small enough so that the latent variables are non-redundant. The goal of the proposed method is to have the number of dimensions (i.e., the number of neural state variables) as close as possible to the degrees of freedom of the observations in the videos. In technical terms, the number of dimensions should be close to the so-called intrinsic dimension (ID), which is the minimum number of independent variables needed to fully describe the state of a dynamical system. Various methods from manifold learning, for instance the one by Levina and Bickel \citep{Levina2004}, are known to efficiently calculate an estimate of the intrinsic dimension. It would be tempting to calculate the intrinsic dimension for the videos and then use it as the bottleneck size of an autoencoder to come up with the latent variables. However, practically, information becomes blurry at much larger bottleneck sizes than the ID already. Therefore, Chen et al. take a two-step approach and define two autoencoders, one regular and one that maps the latent variables of the first to further ID latent variables. These are the neural state variables that can be used for downstream analysis. The approach has not yet been made explainable for scientific discovery.

Generally speaking, neural networks are used in this domain for
\begin{itemize}
    \item making the data sparse in the sense of removing small to negligible interactions \citep{Cranmer2020},
    \item a change of representation (e.g., from coordinates to distances depending on some variables \citep{Cranmer2020}),
    \item data augmentation (to sample arbitrarily large data from the neural network and also smooth the data in that way \citep{Cranmer2020,Li2021}),
    \item the prediction of important parameters to be used in equations directly \citep{Garcon2022}, and
    \item extracting latent variables from low-level input representations (e.g., neural state variables from videos \citep{Chen2022}).
\end{itemize}

\subsection{Representation Learning of the Dynamics and Beyond}

Neural operators \cite{NeuralOperators2023} can learn to map the current state of a system to the next state. This can be done for systems that evolve over space or time and especially for systems for which partial differential equations (PDEs) are too difficult to solve. Neural operators are, however, not restricted to mapping from one state to the next over time: They can learn general functional mappings between various types of inputs and outputs, e.g., inititial conditions to solutions or, even more generally, function-to-function mappings (like DeepONet \cite{KarnadiakisNatMachIntell2021} or Fourier Neural Operators \cite{AnandkumarIJCLR2021}). The latter learn mappings between functions, not just states over time, for instance, they can map a boundary condition (a function) to a solution (another function), which might involve non-temporal variables. Advantages are, amongst others, speed and flexibility (they are not hard to apply from one problem to the next). Neural operators like DeepONet or Fourier Neural Operators are, like other neural networks, black-box models.

\subsection{Open Problems}
Several open problems remain for representation learning of the inputs or learning functional mappings using neural networks:
\begin{itemize}
\item It is currently not well-investigated how learned representations can be aligned with representations that are interpretable by humans.
\item While neural operators can find accurate approximations to the solution of a PDE, understanding how they arrived at that solution is not straightforward. Visualizations, sensitivity analyses, and methods from explainable AI can alleviate some of the problems.
\end{itemize}

\section{Closed-loop Scientific Discovery}
\subsection{Main Concepts, History and Advantages}

\begin{figure}[t!]
    \centering
    \begin{minipage}[b]{0.48\linewidth}
        \centering
        \includegraphics[width=\linewidth]{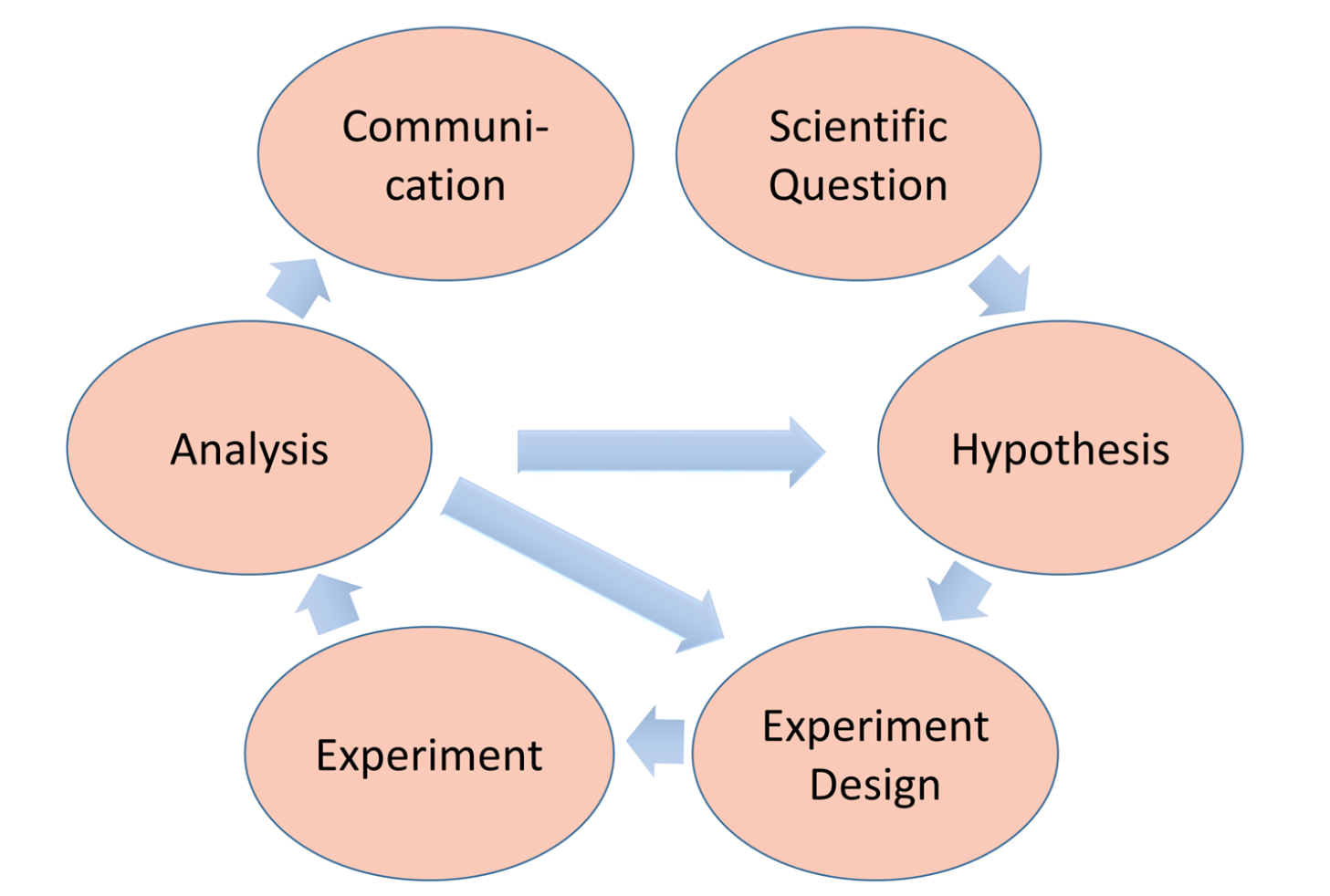}
        \caption{Six steps of the scientific process.}
        \label{fig:six-steps}
    \end{minipage}
    \hfill
    \begin{minipage}[b]{0.48\linewidth}
        \centering
        \includegraphics[width=\linewidth]{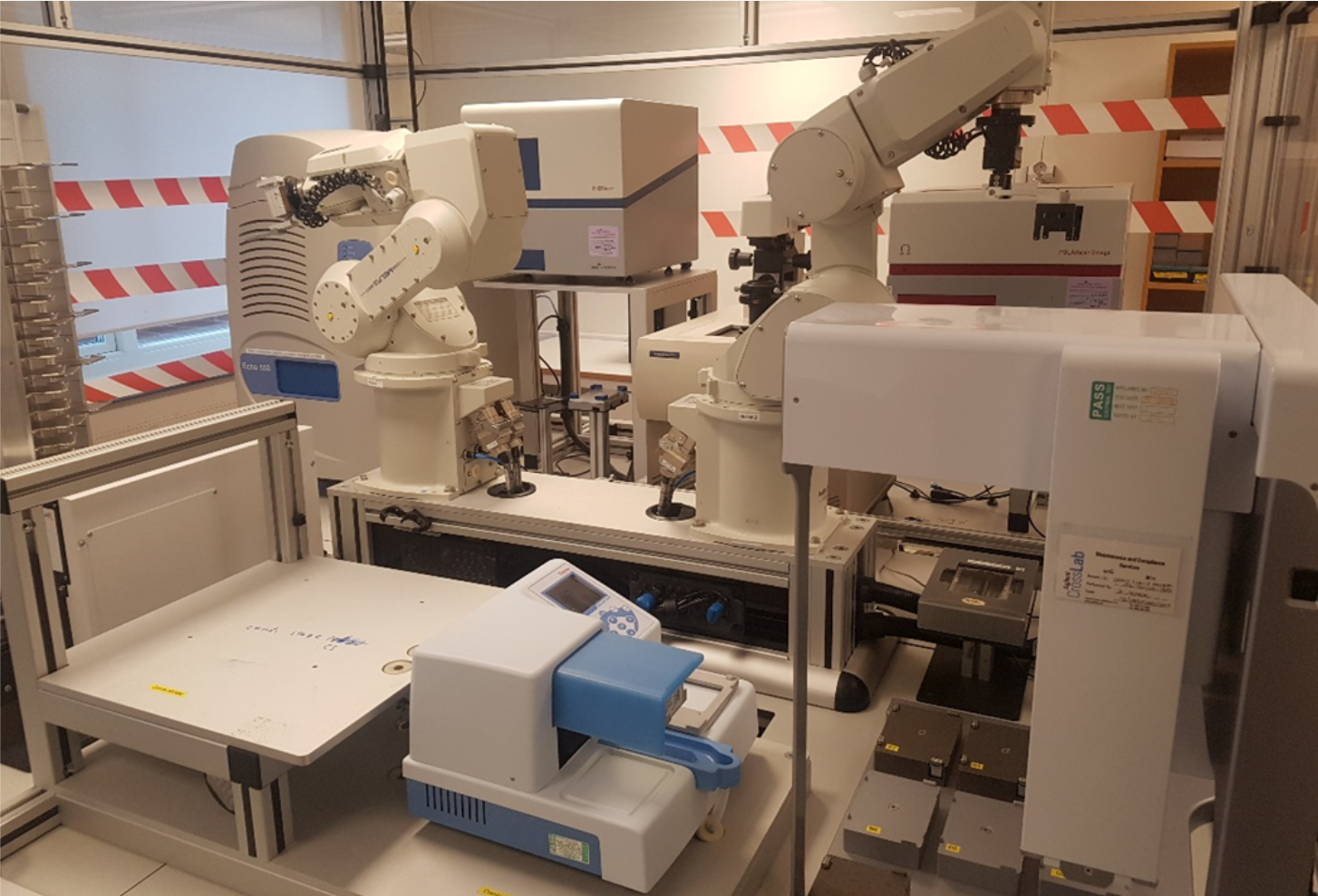}
        \caption{The robot scientist Adam.}
        \label{fig:adam}
    \end{minipage}
\end{figure}

The cutting edge of applying AI to science are ``AI Scientists'' (aka ``Robot Scientists'', ``Self-driving Labs'', ``Autonomous Discovery systems'', ``Machine Scientists'', etc.). These AI systems area capable of the closed-loop automation of scientific research. AI Scientists were named in 2025 by Nature as the ``number one technology to watch''  \cite{eisenstein2025self}. AI Scientists automatically originate hypotheses to explain observations (abduction/induction), devise experiments to test these hypotheses (deduction), physically run the experiments using laboratory robotics, analyze and interpret the results to change the probability of hypotheses, and then repeat the cycle. 
In other words, they aim to automate all or parts of the scientific method, as shown (simplistically) in Figure \Ref{fig:six-steps}. 
As the experiments are conceived and executed automatically by computer, it is possible to completely capture and digitally curate all aspects of the scientific process, making science more reproducible \cite{King2009}.

The first contribution describing a largely autonomous system which discovered new knowledge was due to Ross D. King and his group \citep{King2009}, who developed the Adam robot scientist (see Figure \ref{fig:adam}). Adam identified 6 genes encoding orphan enzymes in yeast (\textit{Saccharomyces cerevisiae}), i.e., enzymes which catalyze reactions occurring in yeast for which the encoding genes were not known at the time. The system was provided with a freezer, liquid handlers, plate readers, robot arms, and further actuators, enabling yeast cultivation experiments lasting as long as 5 days. Yeast growth was measured via optical sensors. On the software side, Adam was provided with an extensive Prolog knowledge base describing known facts about yeast metabolism. Hypotheses were formed by abduction, enabled by a combination of bioinformatic software and databases, after which an experiment planning module was responsible for selecting metabolites to be inserted in the yeast's growth medium.

Another successful example of laboratory automation is Eve. Originally developed for high-throughput drug screening \citep{Sparkes2021}, the system was then instrumental in discovering that several existing drugs could be repurposed to prevent tropical diseases \citep{Williams2015}. Most prominently, it found that an anti-cancer compound (TNP-470) could be employed against the parasite \textit{Plasmodium vivax}, whose bite is the most frequent cause of recurring malaria. The system is able to hypothesize and test quantitative structure--activity relationships (QSARs) via a combination of active learning and Gaussian process regression (GPR). GPR is employed to learn a QSAR $f$ mapping the characteristics of compounds to a response variable indicating the strength of the biological activity; then, the obtained function $f$ is employed as a noisy oracle to select $K$ compounds out of a pool of possible candidates. Exploration and exploitation is balanced. This two-step process may be repeated until enough candidates are obtained. In the meantime, the third generation of robot scientists is being developed.

AI Scientists have a number of relevant advantages, besides being able to discover new knowledge in a way that may be less biased than a human scientist:
\begin{itemize}
\item Efficiency:  AI Scientists are increasing the productivity of science. They can work cheaper, faster, more accurately, and longer than humans \cite{williams2015cheaper}. They can also be easily multiplied.
\item Reproducibility: Biomedical science is facing a ``reproducibility crisis''. AI Scientists have the potential to ameliorate this problem, as they describe experiments in far greater detail and semantic clarity than human scientists, and robots execute experimental protocols more accurately than human scientists \cite{roper2022testing}.
\item Robustness: The Covid-19 pandemic clearly demonstrated the vital importance of biomedical research and the critical need to maintain research continuity \cite{burger2020mobile}.  AI Scientists are increasingly being applied to multiple scientific domains (ranging from quantum mechanics to astronomy, from chemistry to medicine), see Table~\ref{tab:automation_list}. 
\end{itemize}

\subsection{Open Problems}
Three of the current main limitations of AI Scientists are (i) the design of novel experiments, (ii) integration with laboratory robotics, and (iii) the formation of completely new hypotheses and theories.

The central task that faces every experimental scientist is the design of novel experiments to test a hypothesis. The abstract problem is given (1) a hypothesis, and (2) a set of laboratory equipment, output (3) a protocol to test the hypothesis using the equipment. Relatively little AI research has focussed on this aspect of automating science. N.B. that this task is different in kind from the task of traditional ``experimental design'', it also different from deciding, from a set of given experiments, the most efficient (in terms of time/money) to test a set of hypotheses. In all the existing AI Scientists systems that we are aware off the type of experiment that can be executed are limited to a small stereotypical set. For the design of novel experiments to be possible it will be necessary to formalise general scientific knowledge, as well as knowledge about the functionality of laboratory equipment, and experimental protocols. It is also necessary to develop inference and planning engines to generate the new experiments, as well as to develop compilers to translate generated experiments into executable protocols on specific laboratory automation.

\begin{table}[t!]
    \centering
    \begin{tabular}{p{3cm} p{7cm} p{2cm} }
\textbf{Discipline} & \textbf{Name} & \textbf{Country} \\
Drug Discovery & \href{https://www.chalmers.se/en/departments/bio/news/Pages/Chalmers-Robot-Scientist-ready-for-drug-discovery.aspx}{Eve} & Sweden \\
Drug Discovery & \href{https://www.recursion.com/}{Recursion} & US \\
Drug Discovery & \href{https://investor.lilly.com/news-releases/news-release-details/eli-lilly-and-company-collaboration-strateos-inc-launch-remote}{Lilly Life Sciences Studio lab} & US \\
Drug Discovery & \href{https://www.xtalpi.com}{XtaIPi} & China \\
Chemistry & \href{https://www.imperial.ac.uk/rapid-online-analysis-of-reactions}{UK Centre for Rapid Online Analysis of Reactions} & UK \\
Chemistry & \href{https://research.ibm.com/science/ibm-roborxn}{roboRXN at IBM} & Switzerland \\
Chemistry & \href{https://cernaklab.com}{phactor\texttrademark} & US \\
Chemistry & \href{https://jensenlab.mit.edu}{Pharmacy on Demand (PoD)} & US \\
Chemistry & \href{https://moleculemaker.org/}{Molecule Maker Institute} & US \\
Chemistry & \href{https://academic.oup.com/nsr/article/9/10/nwac190/6694008}{AI-Chemist} & China \\
Chemistry & \href{https://www.science.org/doi/10.1126/science.aat0650}{A self-optimizing reactor} & US \\
Chemistry & \href{https://www.chem.gla.ac.uk/cronin/news/cronin-group-builds-chemputer-to-chemify-chemical-space/}{Chemputer} & UK \\
Chemistry & \href{https://www.repository.cam.ac.uk/items/624f2672-bd0b-4dee-a785-783d43854544}{Lapkin Group} & UK \\
Chemistry & \href{https://www.science.org/doi/10.1126/science.adj1817}{RoboChem} & Netherlands \\
Materials & \href{https://www.kebotix.com}{Kebotix} & US \\
Materials & \href{https://www.afrl.af.mil/News/Article/2775183/open-source-software-enables-scientists-to-expedite-research/}{Autonomous Research System (ARES)} & US \\
Materials & \href{https://www.liverpool.ac.uk/leverhulme-research-centre/news/articles/feature-will-we-ever-have-a-robo-chemist/}{Robot Chemist} & UK \\
Materials & \href{https://acceleration.utoronto.ca/}{Acceleration Consortium} & Canada \\
Materials & \href{https://www.bnl.gov/newsroom/news.php?a=116726}{Brookhaven} & US \\
Materials & \href{https://www.chess.cornell.edu/ai-powers-autonomous-materials-discovery}{SARA} & US \\
Materials & \href{https://www.nature.com/articles/s44160-023-00424-1}{AI-Chemist} & China \\
Materials & \href{https://newscenter.lbl.gov/2023/04/17/meet-the-autonomous-lab-of-the-future/}{A-Lab} & US \\
Materials & \href{https://matterhorn.studio/}{Matterhorn} & UK \\
Materials & \href{https://excitonscience.com/}{ARC – Exciton Science} & Australia \\
Materials & \href{https://www.gormleylab.com/}{Gormley} & US \\
Catalysis & \href{https://www.realcat.fr}{RealCat} & France \\
Catalysis & \href{https://swisscatplus.ch/}{SwissCAT+} & Switzerland \\
Metallurgy & \href{https://cordis.europa.eu/project/id/263206}{ACCMET} & EU \\
Materials & \href{https://www.big-map.eu/big-map}{BIG-MAP} & EU \\
Cell Biology & \href{https://www.riken.jp/en/research/labs/bdr/biochem_sim/}{Labdroids} & Japan \\
Cell Biology & \href{https://murphylab.cbd.cmu.edu/}{Murphy Lab} & US \\
Mechanical Eng. & \href{https://www.creativemachineslab.com/}{Creative Machines Lab} & US \\
Protein Design & \href{https://molcure.com/}{Molcure} & Japan \\
Protein Design & \href{https://labgeni.us/}{LabGenius} & UK \\
Systems Biology & Genesis & Sweden \\
Materials/Biology & \href{https://www.anl.gov/autonomous-discovery/developing-a-selfdriving-laboratory-prototype}{Argonne Autonomous Discovery} & US \\
Quantum Physics & \href{https://mariokrenn.wordpress.com/research/}{MELVIN} & Germany \\
Medicine & \href{https://medicine.nus.edu.sg/dbmi/research/automation-science/}{Automation Science} & Singapore \\
  \end{tabular}
  \caption{Robot Scientists by Discipline, Name, and Country}
  \label{tab:automation_list}
\end{table}

Historically, laboratory automation has been driven by the desire to run large numbers of related laboratory experiments, especially in the pharmaceutical and clinical analysis industries. It is now a thriving multibillion dollar industry \cite{king2023robot}. The first use of AI to control laboratory equipment was probably that of Zytkow et al. \cite{ZytkowAAAI1990} (see above). The technology of laboratory automation is steadily advancing, and robots can now carry out most (but not all) of the tasks that humans can do in the laboratory. Such laboratory automation is increasing the productivity of science as robots can work cheaper, faster, more accurately, and for longer (24/7) than humans, they can also be more easily increased/reduced in number. Laboratory automation still has many limitations. Robots typically today operate in protective boxes and are hard to program by bench scientists; logistics tasks are generally performed by lab technicians and scientists, with humans tending the robots for consumables; laboratory automation is expensive in capital to build and maintain - requiring specialised staff. Research in laboratory automation has been largely divorced from AI robotics research - which has mainly focused on the problem of mobile robots. Almost all laboratory robots are fixed in place, although there is growing interest in mobile robot assistants \cite{burger2020mobile}. 

Hypothesis formation needs to be supported by a variety of AI and ML methods, from knowledge representation to active learning and reinforcement learning. The creation of a whole new theory, with theoretical terms and new measurement devices, is at least one level of complexity harder and has not been addressed yet at all.

\section{Autonomy}


\begin{table}[t!]
    \centering
    \caption{Six levels of autonomy in scientific discovery analogously to autonomy levels in autonomous driving}
    \begin{tabular}{|c|p{2cm}|p{5.5cm}|p{3cm}|}
        \hline
        \textbf{Level} & \textbf{Summary} & \textbf{Narrative} & \textbf{Example} \\
        \hline
        0 & No automation & Traditional human science before the advent of computers. & - \\
        \hline
        1 & Machine \ \ \  assistance & The use of computers to automate an aspect of science, e.g. analysing data. & Most current applications of ML. \\
        \hline
        2 & Partial Automation & An important aspect of the discovery cycle is fully automated. & AlphaFold 2, Real-time weather forecasting \\
        \hline
        3 & Conditional Automation & Closed-loop automation. The full cycle of discovery is automated in a restricted domain. & See Table 1. \\
        \hline
        4 & High \ \ \  Automation & Closed-loop automation.   Multiple scientific domains. Limited ability to set its own goals. & No existing system. \\
        \hline
        5 & Full \ \ \ \ \ Automation & All aspects of science are automated and no human intervention is required. & No existing system. \\
        \hline
    \end{tabular}
    \label{tab:automation_levels}
\end{table}


One key aspect of AI Scientists is their degree of autonomy. One approach to measuring autonomy is to use the classification of degrees of autonomy in self-driving cars as \cite{king2023robot}. The approach taken here is similar, Table \ref{tab:automation_levels} describes five levels of autonomy. 

Beyond levels of autonomy are levels of skill. All human drivers are autonomous, but very few are skillful enough drivers to win a Formula 1 race. Among human scientists there are also levels of skill, with few human scientists being skillful enough to win Nobel prize. AI scientists are improving in autonomy and skill. Extrapolating this trend it is likely that advances in technology and our understanding of science will drive the development of ever-smarter AI Scientists. The Physics Nobel Frank Wilczek said that ``in 100 years' time the best physicist will be a machine'' \cite{wilczek2006fantastic}. In February 2020 a workshop was held in London to kick-off the Nobel Turing Grand Challenge to develop: AI systems capable of making Nobel-quality scientific discoveries highly autonomously at a level comparable, and possibly superior, to the best human scientists by 2050 \cite{Kitano2021}. If the Nobel Turing Grand Challenge is achieved this would clearly transform the World, it would be possible to have instead of a few Nobel prize winning ideas a year, hundreds, thousands, millions, ...

\section{Evaluation and Testbeds}
The evaluation of an autonomous discovery system is intrinsically tied to the levels of autonomy displayed by the methodology at hand and which steps of the scientific process are to be automatized and the level of autonomy being evaluated (Figure~\ref{fig:six-steps} and Table~\ref{tab:automation_levels}).
Equation discovery methods may help in automating the analysis of experiments by providing human-readable knowledge, while systems with physical actuators may be evaluated in their ability to execute experimental protocols.
Thus, evaluation methodologies and benchmarks in the area have different characteristics in terms of supervision, data modalities, scope and open-endedness. We define these properties in the following, and give a table of existing methods for evaluation in Table~\ref{tab:testbeds}.

\textbf{Supervision.} 
Supervision refers to the nature of the ground truth or reward signals provided to the autonomous discovery system during training and evaluation. Depending on the degree of autonomy assessed, supervision may range from explicit labels or predefined objectives to feedback signals (rewards in the Reinforcement Learning sense \cite{Sutton2018}). The type and quantity of supervision significantly affect the evaluation outcome, as they directly influence the system's capability to navigate scientific exploration autonomously.

\textbf{Data Modalities}.
Data modalities encompass the types and formats of data available for evaluation, such as pixel-based images, textual descriptions, numerical tables, or structured representations of experimental observations. The choice of modality greatly impacts the complexity and applicability of autonomous systems, as certain data forms inherently require more sophisticated methods for interpretation, abstraction, and knowledge extraction (see Section~\ref{sec:representationlearning}). Evaluating systems across diverse data modalities helps in understanding their flexibility, generalizability, and robustness in real-world scientific scenarios.

\textbf{Scope}.
Scope defines which specific phases of the scientific discovery process the evaluation benchmark addresses. This includes one or more of the six distinct steps: scientific question formulation, hypothesis generation, experimental design, execution of experiments, data analysis and communication.

\textbf{Open-endedness}.
Open-endedness characterizes whether the benchmark or evaluation method includes previously unexplained data, phenomena lacking known mathematical descriptions, or allows the formulation of novel scientific questions. An open-ended benchmark challenges autonomous discovery systems to demonstrate genuine exploratory capabilities, creativity, and adaptability, rather than merely replicating existing knowledge.

We now move to introducing benchmark and testbeds while discussing their potential in the autonomous discovery setting. We will not offer here an exhaustive survey of symbolic regression benchmarks.

\subsection{Available Benchmarks}

\textbf{Nguyen Benchmark Suite}~\cite{nguyen} is a widely-used collection of symbolic regression problems introduced specifically to evaluate genetic programming (GP) methods. It consists of synthetic mathematical equations designed with varying complexity and structure, aiming to assess the ability of GP algorithms to accurately recover symbolic expressions from numerical data. Each task provides numerical input-output pairs generated from known symbolic formulas. The benchmark primarily evaluates one-shot analysis of already collected experimental data.

\textbf{Feynman}~\cite{aifeynman} provides a comprehensive symbolic regression benchmark inspired by fundamental physics equations from the \textit{Feynman Lectures on Physics}. This dataset includes 120 symbolic regression tasks covering a diverse range of physics phenomena, from classical mechanics to electromagnetism.

\textbf{Matbench}~\cite{dunn2020benchmarking} is a supervised machine-learning benchmark containing 13 prediction tasks related to materials science. The dataset consists of structured data representing chemical formulas and crystalline structures, with tasks that involve predicting material properties such as band gap or elastic moduli. It is particularly suited for evaluating analysis capabilities and hypothesis generation for material properties from compositional and structural data. While each task is narrowly defined with a fixed prediction goal, collectively, they support evaluating broad generalizability across material science domains.

\textbf{SCP-116K}~\cite{scp116k} is a large-scale textual dataset comprising problem-solution pairs extracted from higher education science textbooks and other academic sources, totaling 116,000 entries. It is designed primarily for supervised training and evaluation of models on scientific reasoning, question answering, and hypothesis generation from textual data. While each problem-solution pair is relatively constrained in scope, the dataset’s scale and diversity across scientific disciplines provides opportunities for broader generalization and transfer learning evaluation.

\textbf{The Well}~\cite{thewell} is a comprehensive collection of physics simulation datasets, explicitly constructed for machine learning model training and benchmarking in physics-informed learning. It contains diverse simulation data spanning fluid dynamics, astrophysics, plasma physics, and more. These simulations allow evaluation of models’ abilities in hypothesis generation, scientific analysis, and predictive modeling in physics. Its broad diversity and complexity may be employed in open-ended exploration of scientific hypotheses through computational experimentation.

\textbf{ScienceWorld}~\cite{scienceworld2022} is a publicly available reinforcement learning environment designed to evaluate an AI agent's capacity for grounded scientific reasoning in a simulated laboratory context. The benchmark contains 30 interactive text-based tasks, such as converting substances between states of matter. Evaluation relies on binary task completion within limited simulator steps, making it suitable for assessing agents' (abstract, text-based) experimental execution capabilities in a weakly supervised, text-based modality. 

\textbf{DiscoveryWorld}~\cite{jansen2024discoveryworld} is an open-source, highly interactive environment designed to benchmark complete cycles of scientific discovery, including hypothesis generation, experimental design, execution, and analysis. The general setting is akin to a 2D role-playing game to be played on a grid. It provides agents with quests, subquests and various tasks to be completed to make progress. 

\textbf{ChemGymRL}~\cite{chemgym} provides a suite of customizable, publicly accessible reinforcement learning environments simulating chemistry laboratory experiments. Each virtual "bench" simulates distinct chemical procedures such as synthesis or titration. Agents receive structured numeric data representing chemical states and perform sequential lab actions. The library emphasizes experimental design and execution with reward signals, but allows for extension to e.g. new chemical reactions.

\textbf{DiscoveryBench}~\cite{majumder2024discoverybench} is a publicly accessible benchmark focusing on data-driven scientific discovery tasks using multimodal data (tabular data and textual descriptions). It comprises over a thousand real-world and synthetic tasks spanning various scientific domains. Evaluation of agent-generated hypotheses is performed using LLM-based facet analysis, which allows for some open-endedness in the tasks considered. DiscoveryBench primarily targets hypothesis generation and data analysis.

\textbf{BoxingGym}~\cite{gandhi2025boxinggym} provides publicly available, interactive simulation environments for benchmarking autonomous experimental design and scientific model discovery. The benchmark covers multiple scientific domains through generative probabilistic models. Evaluation metrics include expected information gain for experimental quality and predictive power of agent-generated scientific models. The environment is numeric and textual in data modalities and promotes open-ended exploration.

\textbf{Science-Gym}~\cite{cerrato2024sciencegym} is a publicly released Gym-compatible benchmark designed to evaluate autonomous equation discovery in simulated physical and epidemiological environments. Agents interactively select experimental parameters to generate data, subsequently performing symbolic regression to derive underlying scientific equations. Evaluation assesses the symbolic accuracy of discovered equations, providing a structured yet open-ended setting emphasizing experimental execution and analytical reasoning.

\textbf{Open Catalyst 2020 (OC20)}~\cite{chanussot2021open} provides a large-scale benchmark for catalysis research, encompassing over a million atomic structure relaxations generated via density functional theory (DFT) calculations. It offers structured atomic 3D data for supervised machine learning tasks aimed at predicting energies and molecular interactions relevant to catalytic processes. OC20 primarily evaluates data-driven analysis and indirectly supports hypothesis-driven experimental design, particularly aiding in computational screening of catalytic materials. While individually each task has a fixed objective, its expansive dataset encourages robust and generalizable modeling approaches.

\begin{table}[t!]
    \centering
    \caption{Benchmark Categorization by Evaluation Properties. In the \textbf{Scope} column, we take D $=$ experimental Design, E $=$ Experimental Execution, H $=$ Hypothesis formation, A $=$ Analysis of results, Q $=$ research Question formation.\label{tab:testbeds}}
    \small
    \begin{tabular}{lcccc}
        \toprule
        \textbf{Benchmark} & \textbf{Supervision} & \textbf{Data Modalities} & \textbf{Scope} & \textbf{Open-endedness}\\
        \midrule
        Nguyen~\cite{nguyen} & Equation & Tabular & A & No\\
        Feynman~\cite{aifeynman} & Equation & Tabular & A & No\\
        ScienceWorld~\cite{scienceworld2022} & Reward & Text & D, E, A & No\\[0.2em]

        DiscoveryWorld~\cite{jansen2024discoveryworld} & Reward & Text, images & All & Some\\[0.2em]

        ChemGymRL~\cite{chemgym} & Reward & Tabular & E, A & No\\[0.2em]

        DiscoveryBench~\cite{majumder2024discoverybench} & Rewards, LLM judge & Tabular, text & H, A & No\\[0.2em]

        BoxingGym~\cite{gandhi2025boxinggym} & Rewards & Tabular, textual & H, D, E & No\\[0.2em]

        Science-Gym~\cite{cerrato2024sciencegym} & Rewards & Tabular, Images & H, D, E, A & No\\[0.2em]

        Matbench~\cite{dunn2020benchmarking} & Supervised & Tabular & H, A & No\\[0.2em]

        Open Catalyst~\cite{chanussot2021open} & Labels & Tabular, Graph & H, A & No\\[0.2em]

        SCP-116K~\cite{scp116k} & Supervised & Textual & Q, H, A & No\\[0.2em]

        The Well~\cite{thewell} & Equation & Tabular & Q, H, E, A & Yes\\
        \bottomrule
    \end{tabular}
    \label{tab:benchmark_categorization}
\end{table}

\section{Conclusion}

This paper is an attempt at giving a survey of research on automated scientific discovery, from discovering equations to autonomous discovery systems or agents. In doing so, it takes a broad perspective on the topic, which is necessary to understand the individual efforts in context. The article covers the beginnings of the fields to very recent approaches, understanding that we still have a long way of putting everything together to create human-level autonomous scientists. Human-level autonomous scientists should, ultimately, be able to produce whole new theories, along with theoretical terms and measurement devices, which can be communicated to humans and interpreted in the light of other, existing theories. At this point, autonomous discovery systems are focused primarily on ``closing the loop'' and lab automation, and not so much on generating human-interpretable knowledge, like (differential) equations. Vice versa, computational approaches to scientific discovery, e.g., for equation discovery and symbolic regression, do not have the ``embodiment'' in autonomous systems in their focus yet. Ultimately, these currently disparate efforts have to grow together. Finally, it should be noted that artificial intelligence has a role also in so far unexplored areas, like the design of experiments, where much of human ingenuity is currently still needed.

\bibliography{sn-bibliography}

\end{document}